\documentclass{article}

\usepackage{arxiv}

\usepackage[utf8]{inputenc} 
\usepackage[T1]{fontenc}    
\usepackage{hyperref}       
\usepackage{url}            
\usepackage{booktabs}       
\usepackage{amsfonts}       
\usepackage{nicefrac}       
\usepackage{microtype}      
\usepackage{subfig}
\usepackage[table,xcdraw]{xcolor}
\usepackage{graphicx}

\title{COVID-19 Emotion Monitoring as a Tool to Increase Preparedness for Disease Outbreaks in Developing Regions}

\author{
  Santiago Cortes\\
  Factored.ai\\
  Medellin, Antioquia  \\
  \texttt{santiago.cortes@factored.ai} \\ 
\And
 
  Juan Muñoz\\
  Factored.ai\\
  Medellin, Antioquia  \\
  \texttt{juan.manuel.m@factored.ai} \\ 
\And
  David Betancur\\
  EAFIT university\\
  Department of Computer science\\
  Medellin, Antioquia  \\
  \texttt{david.bet.san@gmail.com} \\ 
  \And
 
  Mauricio toro\\
  EAFIT university\\
  Department of Computer science\\
  Medellin, Antioquia  \\
  \texttt{mtorobe@eafit.edu.co} \\
  
}

\begin{document}

\maketitle

\begin{abstract}
The COVID-19 pandemic brought many challenges, from hospital-occupation management to lock-down mental-health repercussions such as anxiety or
depression. In this work, we present a solution for the later problem by developing a Twitter emotion-monitor system based 
on a state-of-the-art natural-language processing model. The system monitors six different emotions on accounts in cities, as well as politicians and health-authorities Twitter accounts.  With an anonymous use of the emotion monitor, health authorities and private health-insurance companies can develop strategies
to tackle problems such as suicide and clinical depression. The model chosen for such a task is a Bidirectional-Encoder Representations from Transformers (BERT) pre-trained on a Spanish corpus (BETO). The model performed well on a  validation dataset. The system is deployed online as part of a web application for simulation and data analysis of COVID-19, in Colombia, available at  \url{https://epidemiologia-matematica.org}.

\end{abstract}

\section{Introduction}

 As a consequence of the COVID-19 pandemic, by April 2020, more than a fifth of the world’s population had already been under some kind of lock-down \cite{lockdownnews}. This type of measure places a burden on mental health; specially, among those who experienced isolation or quarantine \cite{Hossain2020}. In addition, the economic impact caused by the pandemic on developing regions produced a decline on the number of people that could cover their basic needs. Such situation triggered strong emotional responses; ecpecially, among vulnerable populations  \cite{Shammi2020}.

Mental health conditions --such as anxiety and depression-- have been correlated with basic emotions \cite{Power2007}. Emotions are biological states associated with thoughts, feelings, behavioural responses, and a degree of pleasure or displeasure. One of the most used frameworks to categorize emotions is Ekman’s framework \cite{Ekman1992}. According to Ekman, there are six basic emotions: sadness, happiness, fear, anger, surprise and disgust.

Since emotions are correlated with mental health conditions, it is crucial for health authorities and private health-insurance companies to understand how the people's emotions change according to the different lock-down policies implemented during a disease outbreak. To help health authorities, in this task, we want to answer the following question: How can we monitor people’s emotions, related to a disease outbreak, to increase preparedness in developing regions, without posing risks to human rights nor perpetuating inequalities?

There have been some studies on emotion-monitoring of disease outbreaks.  Osherenko implemented lexical-affect sensing, showing the relation of emotions with events that took place in Australia, in 2009 \cite{Osherenko2009}. Recently, Wei et al., gathered a set of 1,690 questions about COVID-19. To classify such questions into 15 categories, using a \emph{Bidirectional Encoder Representations from Transformers (BERT)} which scored 58.1\% accuracy \cite{2005.12522}. To understand emotional responses to lock-downs, Kleiberg et al. asked participants to indicate their emotions and express them in short or long texts \cite{2004.04225}. They also found that by using a predictive model, it was possible to approximate the emotional responses of participants, from text, within 14\% of their actual value. In another work, Wolohan used a \emph{deep Long Short-Term Memory (LSTM}) with \emph{fastText} embeddings to predict population rates of depression, on \textit{Reddit}, in order to estimate the effect of COVID-19 on mental health \cite{Wolohan2020EstimatingTE}. Finally, Kruspe et al. analyzed \emph{Twitter messages (tweets)}, collected during the first months of the COVID-19 pandemic, in Europe, regarding their sentiment \cite{2008.12172}. Kruspe et al. implemented a neural network for sentiment analysis using multilingual-sentence embeddings.

There are two limitations with previous studies in the context of developing regions. First, most of these depend on data and models that only work for the English language. Second, even tough there was a study that considered several languages \cite{2008.12172}, European people may write tweets differently from people in developing regions such as Latin America. For instance, European tweet semantics may be largely based on emoticons and hashtags, which may not be the case in Latin America, and, furthermore, jargon and slang are significantly different.

In this paper, we present a system that continuously access tweets --massively and anonymously-- from accounts in main cities of Colombia, as well as politicians and health-authorities Twitter accounts. The system uses big-data tools such as \emph{Apache Spark} and a public cloud (\emph{Amazon Web Services [AWS]}) to train and deploy a BERT (a state-of-the-art model) that monitors emotions to increase preparedness for disease outbreaks.

\textit{Twitter} is a social network with an enormous amount of public data that can be used to detect specific information of a large sample of the population. Furthermore, information downloaded from Twitter’s free \emph{application programming interface (API)} allows it to capture a large sample of tweets produced in a specific location, while protecting personal information such as gender, race and age, thus the emotion-monitor system does not perpetuate inequalities nor targets specific population groups.

We present methodology in Section 2 and results in Section 3. Finally, we present some conclusions and future work in Section 4.

\section{Materials and Methods}
In this section, we present first data acquisition, pre-processing and training. After, we present model validation.

\subsection{Data Acquisition, pre-processing, filtering and training}
Fine tuning \cite{howard2018universal} and transfer learning \cite{ruder2019transfer} have proven useful to built state-of-the-art models in natural language processing. Nevertheless, the scarcity of labeled data in Spanish was a problem. To overcome the dearth of labeled data, we decided to build our own dataset using Twitter free API and AWS serverless-compute platform \emph{Lambda}. We collected more than 250 Gigabytes of tweets from Colombian principal cities and related accounts, on a time window of one month.  All tweets were labeled using specific words (known as \emph{lexicons}) related with each emotion, proposed by Osherenko \cite{Osherenko2009}.

The tweets were pre-processed removing accentuations and special characters --except for exclamation and question marks. Finally, we removed  the lexicons used to label the tweets and a we used BERT tokenizer. As twitter restricts long texts, the longest text, contained 65 tokens, so we used this size as the maximum-length parameter. 
Tweets were also filtered by considering only those that contains words related to COVID-19 such as coronavirus, epidemics, lock-down, among others.

For training the BETO model, we used a weighted binary cross-entropy loss function, during four epochs, saving a checkpoint on each epoch, to compare the results and use the model with the best results. The optimizer used  was adam with $3 \times 10^{-5}$ as learning rate. 

\subsection{Validation}
Using a \textit{Google-App script}, we made a survey to create a validation dataset to evaluate our model. The web interface showed a random tweet and the user could select one, more than one or no emotions related to the tweet. The users were mainly senior students of Psychology. The selection was saved automatically on a \textit{Google spreadsheet}. A label was saved only when two or more people agreed on it. Finally, we used the labels from the survey to validate the models. The labeled data can be accessed at  \url{https://rb.gy/czggi6}.

To evaluate the model, we considered \emph{mean average precision (map)}, hamming loss and mean F1 score as validation metrics. Hamming loss was used to get an overall result of all predictions, but class weights becomes really important in this problem because there's a lot of joy or sadness tweets but very little that reflect disgust, so f1 and map were evaluated also. We performed four validation experiments as follows:

\begin{enumerate}
  \item The BETO model, with the same pre-process as the training data, but removing the lexicons.
  \item Checking for lexicons associated to an emotion in the tweet and labelling for that emotion.
  \item The BETO model, with the same pre-process as the training algorithm, but without removing the lexicons.
  \item Same as the first one, but using multilingual embeddings and a logistic regression.
\end{enumerate}

Most experiments differ in the pre-processing because the training set was labeled using words included on the tweets and thus, over-fitting was highly probable. To test if the models were learning lexicon presence or actual semantic context, in experiments first and fourth, we decided to perform evaluations on the tweets after removing the lexicons. Finally, to have a baseline mode, in the fourth experiment, we repeated the first experiment,  but with a BERT multilingual embedding.

\begin{figure}[!h]
  \centering
  \includegraphics [width=5in]{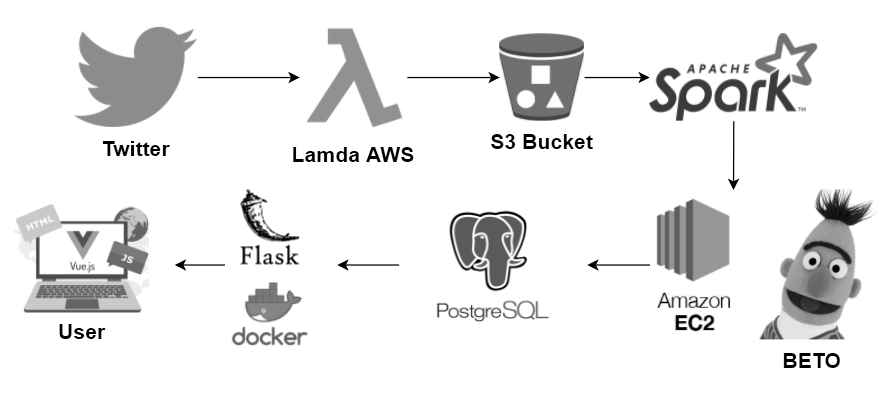}
  \caption{Project deployment architecture on AWS. The system is --currently-- running on AWS.}
  \label{fig:arquitecture}
\end{figure}

The model obtained with \textit{experiment \#1} was trained, validated and deployed in AWS to create a system that is sustainable, resilient and for long-term use, as shown in Figure \ref{fig:arquitecture}.

\section{Results and Discussion}
Table \ref{results-table} shows \emph{map}, hamming loss and mean F1 score of the experiments on the validation dataset. We show the results of the baseline model --using multilingual embeddings and a logistic regression-- in \textit{experiment \#4}. Map and mean F1 are better with higher values, and hamming loss is better with lower values. Highlighted numbers are the best result for each metric. Both map and mean F1 got the best results on  \textit{experiment \#3} and the hamming loss was best on (lexicons-only) \textit{experiment \#2}.

\begin{table}[!htbp]
  \caption{Results with different validation metrics on the validation dataset. }
  \label{results-table}
  \centering
  \begin{tabular}{llll}
    \toprule
    \multicolumn{4}{c}{Validation metrics}                   \\
    \cmidrule(r){2-4}
    \textbf{Experiment}     & \textbf{Mean average Precision}    & \textbf{Hamming loss}    & \textbf{Mean F1} \\
    \midrule
    \textbf{1}              & 0.425          & 0.248             & 0.426 \\
    \textbf{2}              & 0.310          & \textbf{0.152}    & 0.330 \\
    \textbf{3}              & \textbf{0.475} & 0.246             & \textbf{0.450} \\
    \textbf{4}              & 0.320          & 0.165             & 0.200 \\
    \bottomrule
  \end{tabular}
\end{table}


The hamming loss is not class-weight dependent. Therefore, experiment \#2 got the best hamming loss because it is over-fitting on some of the emotions. We found that the tweets are --generally-- more weighted towards joy while toward anger or disgust are rare. Map and mean F1 are more adequate measurements for unbalanced datasets. The BERT models in \textit{experiments \#1} and \textit{\#3} performed well on such unbalanced setting.

\begin{figure}[!htbp]
  \centering
  \includegraphics [width=3in]{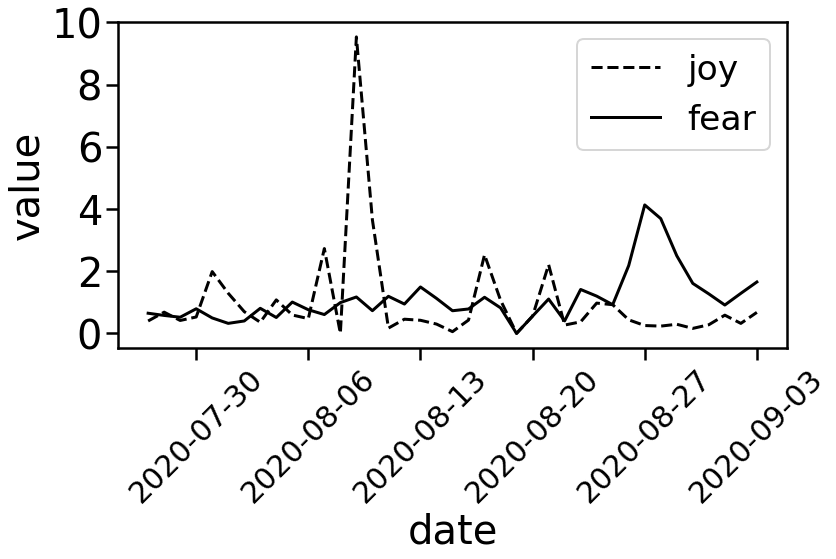}
  \caption{People's emotions on responses to Medellín's major tweets. X-axis represents the day and y-axis the absolute percentage of tweets that belong the emotions of joy or fear. Medellín is the second largest city of Colombia.}
  \label{fig:example_case}
\end{figure}

Figure \ref{fig:example_case} shows the evolution of joy and fear emotions on the people responses to the Twitter account of Medellín's major Daniel Quintero. On August 9th, a peak of joy can be seen: This day the major twitted that he healed from COVID-19. On August 28th, the major announced that the economic reopening was going to happen: This triggered fear among the users reflected on their responses to his tweets.

\section{Conclusions and Future Work}
In this paper, we presented a system that --continuously-- access tweets from accounts in main cities of Colombia as well as from some politicians and health-authorities Twitter accounts in order to train a BERT model that monitors emotions to increase preparedness for disease outbreaks. The deployed system was validated by decision makers of a Colombian private-insurance company.

There are three main conclusions of this work. First, we found out that --oposed to a previous work  \cite{Osherenko2009}--, emoticons and hashtags do not play a significant role in the semantics of tweets in Colombian main cities nor people's responses to politicians and health-authorities Twitter accounts. We believe this is because in developing regions there is not yet a culture of a systematic use of emoticons and hashtags as it is the case of developed regions. Second, the outbreak of COVID-19 exposed the lack of preparedness for such events in many nations around the globe; nonetheless, the tool we presented can be easily extended for future disease outbreaks and other events by simply changing the filtering keywords. Finally, we leave as a future study how to combine Ekman’s six basic emotions to monitor anxiety and depression, as it is proposed in \cite{Power2007}.


\section*{Acknowledgements}
The authors want to thank Juan Pablo Valencia from Colombian private health-insurance company SURA for its support and advice on this research. We also want to thank Juan Jaramillo and Catalina Jaramillo, from Adelphi University, for their comments and suggestions on this research. This research was partially financed by a Minciencias project.

\bibliographystyle{unsrt}  

\end{document}